\documentclass[runningheads]{llncs}
\usepackage{graphicx}

\usepackage{tikz}
\usepackage{comment}
\usepackage{amsmath,amssymb} 
\usepackage{color}

\usepackage[accsupp]{axessibility}  

\begin{document}
\pagestyle{headings}
\mainmatter
\def\ECCVSubNumber{4}  

\title{Visual - Semantic Contrastive Alignment for Few-Shot Image Classification} 

\titlerunning{VS-Alignment}
%
\author{Mohamed Afham, 
Ranga Rodrigo}
\authorrunning{M. Afham et al.}
%
\institute{Dept. of Electronic and Telecommunication Engineering, Univeristy of Moratuwa, \\Sri Lanka \\
\email{afhamaflal9@gmail.com}\\}
\maketitle

\begin{abstract}
Few-Shot learning aims to train and optimize a model that can adapt to unseen visual classes with only a few labeled examples. The existing few-shot learning (FSL) methods, heavily rely only on visual data, thus fail to capture the semantic attributes to learn a more generalized version of the visual concept from very few examples. However, it is a known fact that human visual learning benefits immensely from inputs from multiple modalities such as vision, language, and audio. Inspired by the human learning nature of encapsulating the existing knowledge of a visual category which is in the form of language, we introduce a contrastive alignment mechanism for visual and semantic feature vectors to learn much more generalized visual concepts for few-shot learning. Our method simply adds an auxiliary contrastive learning objective which captures the contextual knowledge of a visual category from a strong textual encoder in addition to the existing training mechanism. Hence, the approach is more generalized and can be plugged into any existing FSL method. The pre-trained semantic feature extractor (learned from a large-scale text corpora) we use in our approach provides a strong contextual prior knowledge to assist FSL. The experimental results done in popular FSL datasets show that our approach is generic in nature and provides a strong boost to the existing FSL baselines.

\end{abstract}

\section{Introduction}

In recent years, deep neural networks have already outperformed humans on image classification with enormous labeled samples supported, which may be against human learning behavior. Humans, however, possess a fast adaptive capacity of recognizing novel classes with a handful of annotated samples. For example, a child can easily generalize the concept of cats and quickly recognize them in reality with only one picture from a book or the Internet. In contrast, existing data-driven deep learning algorithms lag far behind humans in versatility and generalization ability. Therefore, how to construct human-like algorithms and perform visual recognition tasks under data scarcity has important practical value, which also has attracted extensive research interest. To overcome this challenge, few-shot learning (FSL) is introduced for image classification which can learn and generalize from limited data. 

The main paradigm of FSL is training a model on the base classes and requiring it to accurately classify the novel classes with a limited number of examples, which is still threatened by data scarcity. There are various initial line of works study the problem of few-shot learning for image classification \cite{matching,prototypical,maml,Chen2019ACL} and establish strong baselines to improve on top. Meta-learning used to be predominant approach to solve FSL then. However, some recent works adopted standard supervision setting \cite{rfs} along with various self-supervised approaches \cite{skd,boosting,sslfsl} to enhance the quality of the results. However, it is to be noted that visual categories being identified only using class labels (numerical IDs) will seriously limit the contextual features of the category since only a limited number of examples are provided. Identifying this gap, recent line of works \cite{xing2019adaptive,schwartz2019baby,mu2020shaping,rsfsl} adapted using semantic features as a prior knowledge or an auxiliary training mechanism to enhance the FSL performance. RS-FSL \cite{rsfsl} is the recent among all to leverage categorical descriptions to perform few-shot image classifcation. However, it is to be noted that our method utilizes contrastive multimodal alignment for FSL which has never been used in the literature to the best of our knowledge. Further, our approach investigates both visual and semantic attributes in the feature level while RS-FSL predicts the descriptions using the hybrid prototype. The goal of our work is to capture the detailed semantic features and feed it to the visual feature extractor which can then be easily adopted novel categories with very few examples.

In this work we study the effectiveness of contrastive learning which has been proved to perform well \cite{simclr,dino} in standard self-supervised learning. It has also been adapted to multimodal setting as well \cite{clip,avid,crosspoint}. We utilize the simple contrastive learning objective as an auxilliary training mechanism in addition to the standard FSL baseline to provide the contextual knowledge to the model via the semantic prototype generated using a designated semantic feature extractor. We align both the semantic and visual prototypes of each class during an episode of training and employ the contrastive learning learning objective such that the corresponding prototypes regardless of the modalities to be embedded close to each other in the multimodal embedding space. This facilitates a prior knowledge to the visual feature extractor on the semantic attributes of the visual category which is crucial in few-shot image classification.

The major contribution of this approach can be summarized as follows:
\begin{itemize}
    \item We show that a simple contrastive alignment of visual and semantic feature vectors in the embedding space formulates a generalizable visual understanding to perform few-shot image classification.
    \item We introduce an auxiliary contrastive learning objective on top of the existing FSL approach, hence our method is a more generic approach and can be plugged into any of the FSL baselines.
    \item Our experimental results on two standard FSL benchmarks show that multimodal contrastive alignment improves the performance of the standard baselines in FSL problem.
\end{itemize}

\section{Proposed Method}

\begin{figure*}[t]
    \centering
    \includegraphics[width = 0.99 \linewidth]{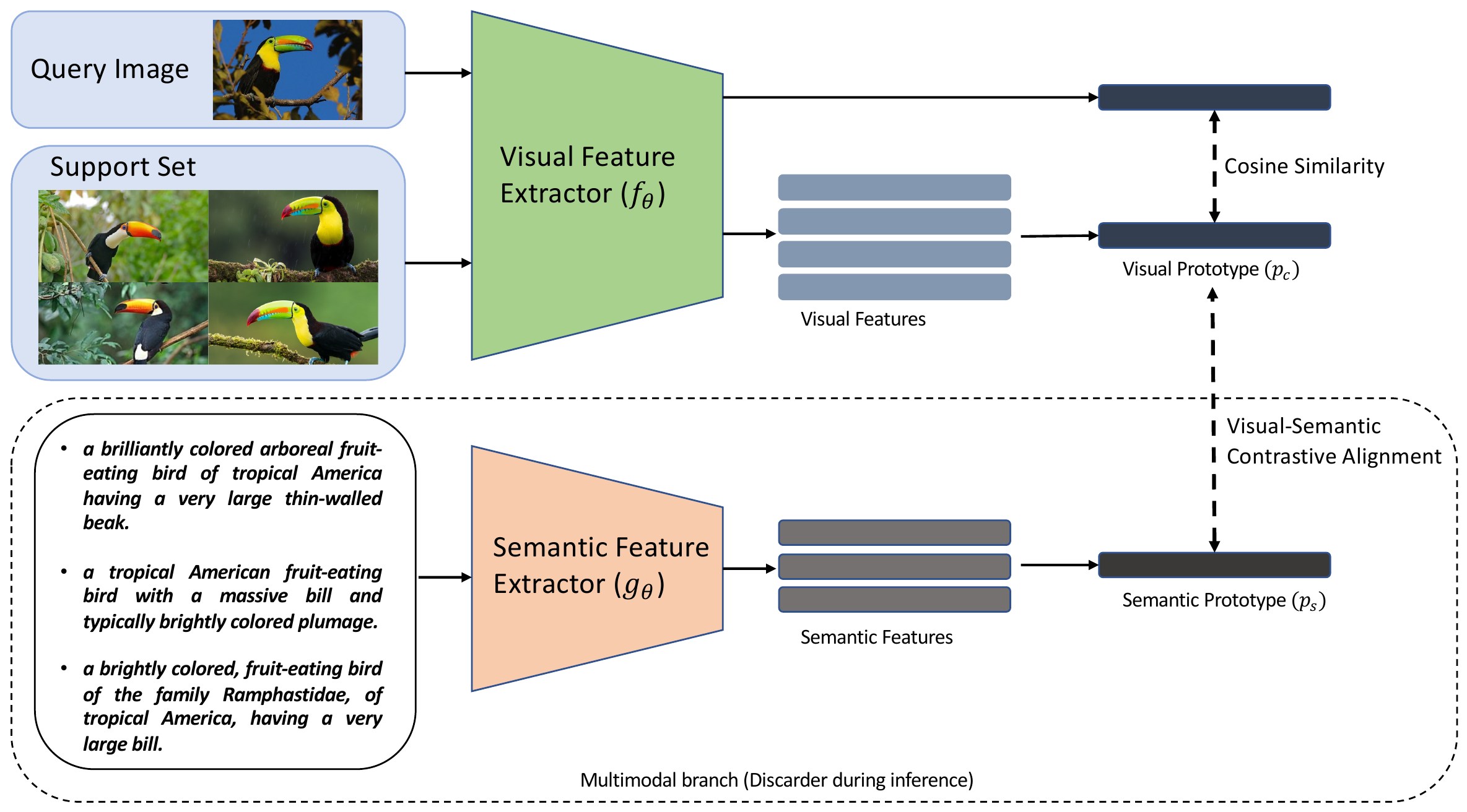}
    \caption{Overall architecture of VS-Alignment for few-shot image classification. Given a support set of images and corresponding class-level descriptions, $f_\theta$ will obtain the visual prototype buy averaging the visual vectors and $g_\theta$ will create the semantic prototype by averaging the semantic feature vectors. During training we employ both the standard cross-entropy (between visual prototype and ground-truth label of query image) and the contrastive alignment (visual and semantic prototypes) as the learning objectives.}
    \label{fig:architecture}
\end{figure*}

In this work, we revisit the contrastive learning objective and leverage it as an auxiliary learning objective in the baseline few-shot learning approach. Given a support set and query images, we introduce an auxiliary contrastive alignment between visual and semantic prototypes to enhance the contextual visual knowledge of the visual prototypes. We utilize Meta-baseline \cite{metabaseline} as our baseline approach and in Sec. \ref{sec:exp} we study the generalizability of our approach with multiple standard FSL baselines. 

This section begins with defining the few-shot learning problem for image classification (Sec. \ref{sec:problem}), followed by explaining about Meta-baseline \cite{metabaseline} for FSL (Sec. \ref{sec:metabaseline}) and finally the descriptions of the proposed add-on architecture to establish the visual-semantic contrastive alignment (Sec. \ref{vsalignment}). We name our approach as VS-Alignment. 

\subsection{Problem Definition} \label{sec:problem}
The standard few-shot image classification paradigm comprises base classes $\mathcal{C}_{base}$, in which there are enough image samples per class and novel classes $\mathcal{C}_{novel}$, where only a limited number of samples are present in each class. The class set between base and novel classes are disjoint i.e., $\mathcal{C}_{base} \cap \mathcal{C}_{novel} = \phi.$ In general, FSL models are trained on K-shot, and N-way episodes. Each episode is created by first sampling N categories from the training set and then sampling two sets of images from these categories: (i) the support set $\mathcal{S}_e = {(s_i,y_i)}_{i=1}^{N \times K}$ containing K examples for each of the N categories and (ii) the query set $\mathcal{Q}_e = {(q_j,y_j)}_{j=1}^Q$ containing Q different examples from the same N categories. After training in this episodic training paradigm, the model is then evaluated in the novel classes $(\mathcal{C}_{novel})$ in the same N-way K-shot setting.

\subsection{Meta-Baseline for FSL} \label{sec:metabaseline}
We adopt the popular and recent baseline named Meta-baseline \cite{metabaseline} to validate our argument of incorporating multimodal contrastive alignment for few-shot image classification enhances the contextual knowledge, hence improving the performance. Prior works to meta-baseline investigated the FSL problem using standard supervision setting \cite{rfs,skd} and episodic learning (meta-learning) setting \cite{matching,prototypical} separately. However, incorporation of both standard supervision and meta-learning arguably produced better results in standard FSL datasets as mentioned by meta-baseline \cite{metabaseline}.

\noindent \textbf{Classification.} During this stage, the model is trained in the base classes $\mathcal{C}_{base}$ in a standard supervision setting. 
Given a dataset of image $(x)$ and label $(y \in \mathcal{C}_{base})$ pairs: 
$\mathcal{D}_{base} = {x_i, y_i},$ the classifier network $f$ maps the input image to a visual feature vector. The visual feature vector is then transformed to the label space to produce the logit $p$ using a linear classifier. This process happens end-to-end and the standard cross-entropy loss is deployed as the learning objective as the following:
\vspace{-0.25em}
\begin{equation}
    \mathcal{L}_{class} = -\text{log} \frac{\text{exp}(p_y)}{\sum_j \text{exp}(p_j)}
\end{equation}

After the classification stage, the last linear classifier layer is removed and the existing embedding module which maps the input image to a visual feature vector is extracted to the meta-learning stage.

\noindent \textbf{Meta-Learning.} During this stage, the episodic learning paradigm has been exploited on top of the supervised trained embedding module. Given a few-shot task with support set $\mathcal{S}_e$, a prototype $p_c$ corresponding to class $c \in \mathcal{C}_{base}$ is computed by averaging the embeddings of all support samples belonging to class $c$:
\vspace{-0.25em}
\begin{equation} \label{eq:prototype}
    p_c = \frac{1}{|\mathcal{S}_e^c|} \sum_{(s_i, y_i) \in \mathcal{S}_e^c} f_\theta(x)
\end{equation}

where $f_\theta$ is the pre-trained visual embedding module. The evaluation is done on the query set $\mathcal{Q}_e$ with the ability of predicting the probability that sample $q_j$ belongs to class $c$ according to the cosine similarity between the embedding of sample $q_j$ and $p_c$: 

\begin{equation}
    p(y = c | q_j, S_e) = \frac{\text{exp}(\tau \cdot \langle f_\theta(q_j), p_c\rangle)}{\sum_k \text{exp}(\tau \cdot \langle f_\theta(q_j), p_k \rangle)}
\end{equation}

Here, $\langle .,. \rangle$ stands for cosine similarity and k ranges for all the classes in the support set of the episode. The learning objective at this stage is a cross-entropy loss computed from $p$ and the labels of the samples in the query- set. During training, each training batch can contain several tasks and the average loss is computed.

\subsection{VS-Alignement for FSL} \label{vsalignment}
With the understanding the potential of contrastive learning which has been immensly deployed in multimodal learning \cite{clip,avid}, we adopted an auxilliary contrastive alignment between the visual and semantic features to enhance the few-shot image classification. To this end, as depicted in Fig. \ref{fig:architecture} we deploy a semantic feature extractor $g_\theta$ which can map the categorical descriptions available to a semantic embedding space as semantic feature vectors. Similar to Eqn. \ref{eq:prototype}, we design a semantic prototype for each class in the support set of the given few-shot episode. We incorporate the proposed multimodal contrastive alignment only on the meta-learning stage of the baseline approach and the classification stage is performed without any modification.

We utilize a transformer model \cite{transformer} similar to what is in CLIP \cite{clip} textual encoder to perform the semantic feature extraction. More details on the implementation will be explained in Sec. \ref{sec:exp}. For each class $c \in \mathcal{C}_{base}$, it is given that we have access to $d_c$ number of categorical descriptions $(w_1, w_2, ..., w_{d_c})$ which we can use for multimodal contrastive alignment. $p_s$ is formulated by averaging the semantic feature vectors of class $c$:
\vspace{-0.5em}
\begin{equation} \label{eq:s_proto}
    p_s = \frac{1}{d_c} \sum_{k = 1}^{d_c} g_\theta(w_k)
\end{equation}

We identify that the visual prototype has the knowledge of the understanding of the given visual dataset, while the semantic prototype is able to contextualize the features since the transformer model \cite{transformer} is able to capture long-range dependencies effectively. Hence, incorporating both the knowledges will yield more generic and fast adoptive understanding of the given visual category. To align both the prototypes, we use simple NT-Xent loss used introduced by Chen \textit{et al.} \cite{simclr}. The auxiliary loss function at this stage to enhace the few-shot image classification is defined as:

\begin{equation}
\footnotesize
    \mathcal{L}_{vs}(i, p_c, p_s) = -\log \frac{\exp(\langle p_{c_i}, p_{s_i} \rangle/\tau)}{ \sum\limits_{\substack{k=1 \\ k \neq i}}^{N} \exp(\langle p_{c_i}, p_{c_k} \rangle/\tau) + \sum\limits_{\substack{k=1}}^{N} \exp(\langle p_{c_i}, p_{s_k} \rangle/\tau)}
\end{equation}

The total learning objective is defined as the weighted combination of both the visual learning objective and the multimodal learning objective:

\begin{equation}
    \mathcal{L} = \mathcal{L}_{class} + \lambda \mathcal{L}_{vs}
\end{equation}

Here, $\lambda$ is a weighting factor and is a tunable hyperparameter determined using grid-search.  

\section{Experiments} \label{sec:exp}
\noindent \textbf{Datasets.} We conduct experiments on two benchmark datasets for few-shot image classification: mini-ImageNet \cite{matching} and CUB \cite{cub}. The miniImageNet dataset consists of 100 image classes extracted from the original ImageNet dataset \cite{imagenet}. Each class contains 600 images of size $84 \times 84$. We follow the splitting protocol proposed by \cite{prototypical}, and use 64 classes for training, 16 for validation, and 20 for testing. We obtained the categorical descriptions provided by \cite{rsfsl}.

The CUB dataset contains 200 classes and 11 788 images in total. We split the dataset into 100 classes for training, 50 for validation, and 50 for testing following the prior standard works \cite{sslfsl,rsfsl}. The categorical description for CUB is obtained from \cite{cublang}. We randomly sample the required number of descriptions. 

\noindent \textbf{Implementation Details.} To be in fair comparison with the existing works, we deploy the 4-layer convolutional architecture proposed in \cite{prototypical} for CUB and ResNet-12 \cite{resnet} for miniImageNet. For semantic feature extractor we use pre-trained textual encoder trained using CLIP \cite{clip} model in all of our experiments. The model comprises of 12-layer transformer model \cite{transformer} with 8 attention heads and 512-width. Following \cite{skd}, during classification stage, we use SGD optimizer with an initial learning rate of 0.05, momentum of 0.9, and weight decay of 0.0005. We train the model for 100 epochs with a batch size of 64 and the learning rate decays twice by a factor of 0.1 at 60 and 80 epochs.
During the vs-alignment (meta-learning stage), we use a contant learning rate of 0.001 with Adam optimizer and train the model for 600 epochs. We define $\lambda = 2.5$ based on the grid-search we performed. All the experiments were performed using nvidia Quadro RTX 6000 single-GPU and we report the results form 5-way 1-shot setting.

\begin{table}[t]
\centering
\begin{tabular}{c c c c} 
 \hline
\textbf{Method} & \textbf{Backbone} & \textbf{Accuracy} \\ [0.5ex] 
 \hline
 MatchingNet \cite{matching} & Conv-4 & 60.52$\pm$0.88 \\ 
 MAML \cite{maml} & Conv-4 & 54.73$\pm$0.97 \\
 ProtoNet \cite{prototypical} & Conv-4 & 50.46$\pm$0.88 \\ 
 RFS \cite{rfs} & Conv-4 & 41.47$\pm$0.72 \\ 
 L3 \cite{andreas2017learning} & Conv-4 & 53.96$\pm$1.06  \\  
 LSL \cite{mu2020shaping} & Conv-4 & 61.24$\pm$0.96 \\ 
 Chen \textit{et al.} \cite{Chen2019ACL} & Conv-4 & 60.53$\pm$0.83 \\ 
 Meta-Baseline \cite{metabaseline}  & Conv-4 & 59.30$\pm$0.86 \\
 RS-FSL \cite{rsfsl} & Conv-4 & 65.66$\pm$0.90 \\ 
 \hline
 VS-Alignment  & Conv-4 & \textbf{66.73$\pm$0.78} \\
 \hline
\end{tabular}
\caption{Performance comparison  on the CUB dataset. We report average 5-way 1-shot accuracy (\%) with 95\% confidence interval. Table is an extended version adapted from RS-FSL \cite{rsfsl}.
}
\label{table:SOTA_CUB}
\vspace{-2em}
\end{table}

\subsection{Comparison with the baselines}

We report the results of our approach along with the comparison of the state-of-the art results in CUB dataset in Tab. \ref{table:SOTA_CUB}. It is clear that the proposed visual-semantic alignment method outperforms the baseline approach and some of the existing state of the art approaches. This shows the importance of incorporating the semantic knowledge of the few-shot category in a contrastive style. The results on miniImagenet dataset is reported in Tab. \ref{table:SOTA_miniimagenet}. It is to be noted that even though there is an improvement in performance over the baseline \cite{metabaseline} in the proposed approach, the gap is not significant. We hypothesize that it could have happened because of the lack of categorical descriptions compared to that of CUB dataset. In both of the experiments, we compare our approach with RS-FSL \cite{rsfsl} which is a recent FSL approach that utilizes categorical descriptions as well. 
\vspace{-1em}

\begin{table}[h]
\centering
\begin{tabular}{c c c c} 
  \hline
  \textbf{Method} & \textbf{Backbone} & \textbf{Accuracy} \\ [0.5ex] 
 \hline
 ProtoNet \cite{prototypical} & Conv-4 & 55.50$\pm$0.70 \\ 
 Matching Net \cite{matching}& Conv-4 & 43.56$\pm$0.78 \\ 
 MAML\cite{maml} & Conv-4 & 48.70$\pm$1.84 \\ 
 Chen \textit{et al.} \cite{Chen2019ACL} & Conv-4 & 48.24$\pm$0.75 \\ 
 Boosting \cite{boosting}& WRN-28-10 & 63.77$\pm$0.45\\ 
 RFS-Simple \cite{rfs} & ResNet-12 & 62.02$\pm$0.63\\ 
 RFS-Distill \cite{rfs} & ResNet-12 & 64.82$\pm$0.60 \\ 
Meta-Baseline \cite{metabaseline} & ResNet-12 & 63.17$\pm$0.23 \\ 
RS-FSL \cite{rsfsl}  & ResNet-12 & 65.33$\pm$0.83 \\
 \hline
 VS-Alignment  & ResNet-12 & \textbf{65.89$\pm$0.80} \\
 \hline
\end{tabular}
\caption{Comparison with prior works on the miniImageNet. 
}
\label{table:SOTA_miniimagenet}
\vspace{-2em}
\end{table}

Tab. \ref{table:ablation_different_baselines} reports the results of FSL across multiple standard baselines. It is descriptive that the addition of visual-semantic alignment boosts the performance in both ProtoNet \cite{prototypical} and Meta-baseline \cite{metabaseline} while it depreciates the performance in RFS and SKD \cite{rfs,skd}. Since RFS and SKD are of non-episodic FSL methods, we come to a conclusion that we can plug in pur method and boost the performance only episodic few-shot learning paradigm.

\begin{table}[h]
\centering\setlength{\tabcolsep}{4pt}
\scalebox{0.85}{
\begin{tabular}{c c c c} 
 \hline
 \textbf{Baseline} & \textbf{Backbone} & \begin{tabular}{l}\textbf{Without}\\ \textbf{VS-Alignment}\end{tabular} & \begin{tabular}{l}\textbf{With}\\ \textbf{VS-Alignment}\end{tabular}\\
 \hline
 ProtoNet \cite{prototypical}  & Conv-4 & 57.97$\pm$0.96 & 61.43±0.83 \\ 
 RFS \cite{rfs} & Conv-4 & 44.93$\pm$0.76 & 42.36±0.64 \\
 SKD \cite{skd} & Conv-4 & 58.75$\pm$0.96 & 56.43±0.43 \\
 Meta-Baseline \cite{metabaseline} & Conv-4 & 59.30±0.86 & \textbf{66.73±0.78} \\
 \hline
\end{tabular}}
\caption{Performance of different baselines both with and without the proposed visual-semantic contrastive alignment on the CUB dataset. 
}
\label{table:ablation_different_baselines}
\end{table}
\vspace{-4em}
\section{Conclusion}
In this work, we introduce a simple contrastive alignment between visual and semantic prototypes of visual categories which acts as an auxiliary task to faciliate few-shot image classification. Our approach is generic in nature and can be plugged into any meta-learning based few-shot baselines. We also prove that our approach outperforms multiple standard baselines in the 5-way 1-shot few-shot setting hence establishing a new research direction to solve few-shot image classification task.

\bibliographystyle{splncs04}
\bibliography{egbib}
\end{document}